\documentclass[preprint,12pt,3p]{elsarticle}

\usepackage{graphics}
\usepackage{graphicx}
\usepackage{epsfig}
\usepackage{xcolor}

\usepackage{amssymb}
\usepackage{amsthm}
\usepackage{tablefootnote}
\usepackage{lineno}
\usepackage{amsmath,amssymb} % define this before the line numbering.
\usepackage{color}
\usepackage{algorithm}
\usepackage{algorithmic}
\usepackage{subfig}
\usepackage{amsmath}
\usepackage{array,multirow}
\usepackage{makecell}

\newcounter{inlineenum}
\renewcommand{\theinlineenum}{\alph{inlineenum}}

\journal{Microprocessors and Microsystems Journal}

\begin{document}

\begin{frontmatter}
%%%%%%%%%%%%%%%%%%%%%%%%%%%%%%%%%%%%%%%%%%%%%%%%%%%%%%%%%
%%%%%%%%%%%%%%%%%%% Title %%%%%%%%%%%%%%%%%%%%%%%%%%%%%%%
%%%%%%%%%%%%%%%%%%%%%%%%%%%%%%%%%%%%%%%%%%%%%%%%%%%%%%%%%
\title{A Fully Pipelined FPGA Accelerator for Scale Invariant Feature Transform Keypoint Descriptor Matching}
%%%%%%%%%%%%%%%%%%%%%%%%%%%%%%%%%%%%%%%%%%%%%%%%%%%%%%%%%

%%%%%%%%%%%%%%%%%%%%%%%%%%%%%%%%%%%%%%%%%%%%%%%%%%%%%%%%%
%%%%%%%%%%%%%%%%%%% Authors %%%%%%%%%%%%%%%%%%%%%%%%%%%%%
%%%%%%%%%%%%%%%%%%%%%%%%%%%%%%%%%%%%%%%%%%%%%%%%%%%%%%%%%
\author[]{Luka Daoud}
\ead{LukaDaoud@u.boisestate.edu}

\author[]{Muhammad Kamran Latif}
\ead{MuhammadLatif@u.boisestate.edu}

\author[]{H S. Jacinto}
\ead{SheltonJacinto@u.boisestate.edu}

\author[]{Nader Rafla}
\ead{nrafla@boisestate.edu}
%%%%%%%%%%%%%%%%%%%%%%%%%%%%%%%%%%%%%%%%%%%%%%%%%%%%%%%%%

%%%%%%%%%%%%%%%%%%%%%%%%%%%%%%%%%%%%%%%%%%%%%%%%%%%%%%%%%
%%%%%%%%%%%%%%%%%%% Affiliation %%%%%%%%%%%%%%%%%%%%%%%%%
%%%%%%%%%%%%%%%%%%%%%%%%%%%%%%%%%%%%%%%%%%%%%%%%%%%%%%%%%
\address{Department of Electrical and Computer Engineering \\ Boise State University \\ Boise, ID 83725, USA}
%%%%%%%%%%%%%%%%%%%%%%%%%%%%%%%%%%%%%%%%%%%%%%%%%%%%%%%%%

%%%%%%%%%%%%%%%%%%%%%%%%%%%%%%%%%%%%%%%%%%%%%%%%%%%%%%%%%
%%%%%%%%%%%%%%%%%%% Abstract %%%%%%%%%%%%%%%%%%%%%%%%%%%%
%%%%%%%%%%%%%%%%%%%%%%%%%%%%%%%%%%%%%%%%%%%%%%%%%%%%%%%%%
\begin{abstract}
The scale invariant feature transform (SIFT) algorithm is considered a classical feature extraction algorithm within the field of computer vision. The SIFT keypoint descriptor matching is a computationally intensive process due to the amount of data consumed. In this paper, we designed a fully pipelined hardware accelerator architecture for the SIFT keypoint descriptor matching. It was implemented and tested on a field programmable gate array (FPGA). The proposed hardware architecture is able to properly handle the memory bandwidth necessary for a fully-pipelined implementation and hit the roofline performance model achieving the potential maximum throughput. The fully pipelined matching architecture was designed based on consine angle distance approach. It was optimized for 16-bit fixed-point operations and implemented on hardware using Xilinx Zynq-based FPGA development board. Our proposed architecture showed a noticeable reduction of area resources compared with its counterparts in the literature maintaining high throughput by alleviating the memory bandwidth restrictions. The results showed reduction in device-resources up to 91\% in LUTs and  79\% of BRAMs. Our hardware implementation is 15.7$\times$ faster than the comparable software approach.
\end{abstract}
%%%%%%%%%%%%%%%%%%%%%%%%%%%%%%%%%%%%%%%%%%%%%%%%%%%%%%%%%

%%%%%%%%%%%%%%%%%%%%%%%%%%%%%%%%%%%%%%%%%%%%%%%%%%%%%%%%%
%%%%%%%%%%%%%%%%%%% Keywords %%%%%%%%%%%%%%%%%%%%%%%%%%%%
%%%%%%%%%%%%%%%%%%%%%%%%%%%%%%%%%%%%%%%%%%%%%%%%%%%%%%%%%
\begin{keyword}
Scale Invariant Feature Transform \sep  SIFT \sep  Matching algorithm \sep  FPGA \sep   Pipeline \sep  Acceleration \sep  High Level Synthesis \sep  HLS.
\end{keyword}
%%%%%%%%%%%%%%%%%%%%%%%%%%%%%%%%%%%%%%%%%%%%%%%%%%%%%%%%%
\end{frontmatter}
%%%%%%%%%%%%%%%%%%%%%%%%%%%%%%%%%%%%%%%%%%%%%%%%%%%%%%%%%\linenumbers

%%%%%%%%%%%%%%%%%%%%%%%%%%%%%%%%%%%%%%%%%%%%%%%%%%%%%%%%%
%%%%%%%%%%%%%%%%%%% Introduction %%%%%%%%%%%%%%%%%%%%%%%%
%%%%%%%%%%%%%%%%%%%%%%%%%%%%%%%%%%%%%%%%%%%%%%%%%%%%%%%%%
\section{Introduction}
Object recognition using feature-based algorithms are generally computationally intensive. The scale-invariant feature transform (SIFT) algorithm proposed in 1999 by David Lowe \cite{DavidLowe1999}, is a classical and well-known algorithm within the field of computer vision. SIFT algorithm is a feature-based algorithm that can be applied in object recognition. The best candidate match for a SIFT keypoint is found by identifying its nearest-neighbor in the keypoint database. The matching process often involves operating on data-at-rest but more recently real-time applications using feature-based object recognition have gained popularity. Feature extraction based object recognition is an approach commonly applied in several varying applications such as medical imaging \cite{MedApp1}, satellite imaging \cite{SatelliteApp2}, facial recognition \cite{FacialRecog1}, and the landing of unmanned aerial vehicles (UAVs) \cite{UVApp1}.

Various steps in the extracting SIFT descriptors often require the use of complex software routines that require intensive computations \cite{DavidLowe1999}. However, in a running scenario of keypoint extraction, the extraction only occurs once per test image. The limitations of keypoint descriptor matching thus requires that matching must be performed every time a test image is compared with a possible match in the database. Each time the database needs to be accessed, the overall matching time for the test image increases as the overall size of the database grows.

The SIFT descriptor matching is based on the nearest-neighbor algorithm \cite{DavidLowe1999} where, for a single test keypoint descriptor match, the Euclidean distances \cite{kolman2004elementary} of the test descriptor are calculated between each descriptor in the descriptor database. The  calculated distances are then sorted such that the minimum and second minimum distances are found. A positive match between the test descriptor and the descriptor database is found if the Euclidean distance ratio is above a pre-set threshold, suggested by David Lowe in \cite{DavidLowe1999}.

Since a SIFT keypoint descriptor is an array of 128 elements, calculated based on all pixels of an image around the centered keypoint in a 16$\times$16 sliding window. The generated descriptor by this method can be defined mathematically as:
\[ d_{k}^{\alpha} = \{ f_{k,1}^{\alpha}\ ,\ f_{k,2}^{\alpha}\ ,\ \dots \dots\ ,\ f_{k,128}^{\alpha} \} \ \ .\] 
The Euclidean distance between two descriptors, $d_{k}^{\alpha}$ and $d_{m}^{\beta}$, is thus calculated:
\[ 
    \sum\limits_{i=1}^{128}
        \frac{( f_{k,i}^{\alpha} - f_{m,i}^{\beta})^{2}}
             {(f_{k,i}^{\alpha} + f_{m,i}^{\beta})} 
         .\]
In the process of matching a descriptor, $d_{k}^{\alpha}$, with a database, the Euclidean distances of $d_{k}^{\alpha}$ in relation to the database's descriptors is calculated. The process of calculating Euclidean distances is computationally intensive however, resource consumption can effectively be reduced by changing the calculation of Euclidean distance. Instead of using a conservative approach of calculating the Euclidean distance as mentioned, a cosine angle distances can be calculated between the descriptors \cite{Chi_square}. Since SIFT descriptors are normalized during keypoint extraction, calculating the angular distances by taking the arc-cosine of the dot-products of normalized descriptors prove to be a close approximation for Euclidean distances \cite{Chi_square}. Utilizing a method of angular distance will significantly reduce the hardware resource consumption.

If an image of $m$ descriptors is represented by a matrix of size $m\times128$, there is a recurrent redundancy of memory access for descriptors of an image and the descriptor database. Memory access times for descriptors further vary based on locality thus, in a software approach, memory access time becomes variant that may impact on both timing and resource overheads for the SIFT descriptors matching.

In this paper, our proposed SIFT descriptors matching architecture is designed and implemented with the purpose of accelerating the matching process, handling the memory bandwidth limitation, and reducing the area resources. The contributions of this paper are summarized as following:
\begin{itemize}
    \item A hardware implementation of the SIFT keypoint descriptor matching based on cosine angle distance on FPGA including:
    \begin{itemize}
        \item A fully pipelined architecture.
        \item Minimal resource utilization.
        \item High throughput hardware accelerator.
    \end{itemize}
    \item Resulting analysis of memory bandwidth usage and its effect on the overall computational performance.
\end{itemize}

The rest of this paper is organized as follows: Section \ref{sec:related_work} summarizes the  literature review and the related work of SIFT descriptors matching on accelerating platforms. Section \ref{sec:background} provides background and related definitions along with the software approach of the matching algorithm based on calculating cosine angle distances. Section \ref{sec:comp_mem_BW} studies the computation and memory bandwidth optimization. Section \ref{sec:proposed_HW} presents our proposed matching architecture on FPGA. Section \ref{sec:exp_eval} evaluates our proposed matching architecture and provides the experimental results. Finally, Section \ref{sec:conclusion} concludes the paper.
%%%%%%%%%%%%%%%%%%%%%%%%%%%%%%%%%%%%%%%%%%%%%%%%%%%%%%%%%

%%%%%%%%%%%%%%%%%%%%%%%%%%%%%%%%%%%%%%%%%%%%%%%%%%%%%%%%%
%%%%%%%%%%%%%%%%%%% Section Related Work %%%%%%%%%%%%%%%%
%%%%%%%%%%%%%%%%%%%%%%%%%%%%%%%%%%%%%%%%%%%%%%%%%%%%%%%%%
\section{Related Work}\label{sec:related_work}
This section particularly focuses on different approaches of calculating nearest-neighbor distances for descriptors matching algorithms on FPGA-based accelerators. It also provides a brief overview of the matching implementations on other platforms. 

There have been several hardware-based implementations of descriptors matching on FPGA \cite{Vourvoulakis, comparisonPaper2017, Wang, KAPELA}. Most recently, Vourvoulakis et~al. \cite{Vourvoulakis} proposed an FPGA-based architecture for SIFT descriptors matching based on the calculation of the distances between the descriptors in the database. The similarity between the descriptors was determined based on the minimum value of SAD (Sum of Absolute Distances) calculators. Their implementation was based on comparing the currently extracted descriptor with 128 previously detected ones to find a potential match. The authors proposed a moving window of 16 descriptors to fit the entire matching architecture on an FPGA. In their implementation, a total 8 clock-cycles were required to calculate 128 SAD values to report a potential match using a single matching core that required significant memory resources.

Lentaris et~al. \cite{comparisonPaper2017} implemented a pipelined architecture for SIFT descriptor matching using the Euclidean norm for computing distances between descriptors. In their implementation, a finite state machine fetches all the descriptors from the test image $d_{k,i}^{\alpha}$ and the descriptors from the database $d_{k,i}^{\beta}$ stored in memory one by one. The descriptor pair is passed to a chi-square distance state, where the similarity of the two descriptors was evaluated by calculating the distance between them. The distance calculating state consists of 128 chi-square ($\chi^2$) calculators and each calculator performs $(d_{k,i}^{\alpha} - d_{k,i}^{\beta})^2/(d_{k,i}^{\alpha} + d_{k,i}^{\beta})$ calculation where $i$ is the $i^{th}$ element of the 128-dimensional vector. Each multiplier and divider in chi-square state is 16-bit and produces a 32-bit result. The output from $\chi^2$ calculators is summed using linear systolic array and the result is passed to matching state to keep tracking of the two best matches. At the end of the database, the distance ratio of these two matches is compared with a fixed threshold to accept or reject the best match. Their used technique of the matching algorithm by calculating the Euclidean distance necessitated more resources than our approach as explained in Section \ref{sec:exp_eval}.

Wang et~al. \cite{Wang} proposed an embedded System-on-Chip for features detection and matching. Their system extracts binary robust independent elementary features (BRIEF) \cite{BRIEF} descriptors from the detected SIFT ones. Unlike SIFT descriptors that has 128 elements, the BRIEF descriptor is a vector of 64 elements. The BRIEF matching detection was performed by calculating the distances between two BRIEF descriptors. A successful match is reported if the calculated distance is smaller than a minimum threshold \cite{Wang}. 

Kapela et~al. \cite{KAPELA} presented a hardware-software platform in which fast retina keypoint (FREAK) \cite{FREAK} descriptors were extracted in software and matched by calculating the Hamming distance which was implemented on Xilinx Zynq-7000 FPGA. Their proposed matching core included multiple Hamming distance calculator circuits that are running in parallel to calculate the distance between the descriptors. The overall performance of their system depends on the number of the Hamming distance cores. Additionally, the number of LUTs and registers increases proportionally with the number of Hamming calculators.

Condello et~al. \cite{CONDELLO} presented an OpenCL-based feature matching algorithm that made use of the capabilities of GPUs to speedup the matching process for speeded-up robust feature (SURF) descriptors. The matching algorithm uses Euclidean distances to calculate the nearest neighbors for a test descriptors with the others in the database. They implemented their matching core on NVIDIA's GTX275, which has a theoretical peak of 2760 GFlops. However, the latency of the global memory access affected on the computation power of the GPU where it limited the memory reuse during the distance computation step of the matching process.

Fassold et~al. \cite{fassold2015gpu} used NVIDIA's Tesla K20 GPU for the SIFT descriptor matching by calculating the nearest-neighbors between descriptors using Euclidean distance. Their implementation of the matching architecture on the GPU achieved 13 milliseconds for a set of 2,800 descriptors.

The matching algorithm for most of the implementations is based on calculating the nearest-neighbor distances between the current feature and the features in the database. To the best of our knowledge, this paper is the first attempt for hardware implementation of SIFT matching algorithm on FPGA, where the matching technique is based on calculating the nearest-neighbor distances using cosine angle distance rather than using the traditional descriptor distance calculations. The following part of this paper moves on to describe in greater detail the SIFT matching algorithm based on cosine angle distance technique.
%%%%%%%%%%%%%%%%%%%%%%%%%%%%%%%%%%%%%%%%%%%%%%%%%%%%%%%%%

%%%%%%%%%%%%%%%%%%%%%%%%%%%%%%%%%%%%%%%%%%%%%%%%%%%%%%%%%
%%%%%%%%%%%%%%%%%%% Section %%%%%%%%%%%%%%%%%%%%%%%%%%%%%
%%%%%%%%%%%%%%%%%%%%%%%%%%%%%%%%%%%%%%%%%%%%%%%%%%%%%%%%%
\section{Matching Algorithm based on Cosine Angle Distance} \label{sec:background}

\subsection{Nomenclature and Definitions}

An image is a 2-D array of pixels that carry information and keypoint descriptors are highly distinctive features in an image. A SIFT descriptor is a vector of 128 elements that describe a scale-invariant local image region. It can be given as $d_{k}^{\alpha}$, where $k$ and $\alpha$ are the $k^{th}$ descriptor in an image $\Tilde{\alpha}$.
\[ d_{k}^{\alpha} = \{ f_{k,1}^{\alpha}\ ,\ f_{k,2}^{\alpha}\ ,\ \dots \dots\ ,\ f_{k,128}^{\alpha} \},\] where $f_{k,i}^{\alpha}$ is the $i^{th}$ element of the $k^{th}$ descriptor of image $\Tilde{\alpha}$ and $(0 \leq\ f_{k,i}^{\alpha} \leq 1)$. So, descriptors of an image $\Tilde{\alpha}$ that has a $m$ set of descriptors is described as $d^{\alpha}$:
\[ d^{\alpha} = 
\begin{bmatrix}
    d_{1}^{\alpha} \\ 
     d_{2}^{\alpha} \\ 
    \vdots \\ 
    d_{m}^{\alpha}
\end{bmatrix}=
\begin{bmatrix}
    f_{1,1}^{\alpha} & f_{1,2}^{\alpha} & f_{1,3}^{\alpha} & \dots  & f_{1,128}^{\alpha} \\
    f_{2,1}^{\alpha} & f_{2,2}^{\alpha} & f_{2,3}^{\alpha} & \dots  & f_{2,128}^{\alpha} \\
    \vdots & \vdots & \vdots & \ddots & \vdots \\
    f_{m,1}^{\alpha} & f_{m,2}^{\alpha} & f_{m,3}^{\alpha} & \dots  & f_{m,128}^{\alpha}
\end{bmatrix}
.\]
The dot-product operation of two descriptors, $d_{l}^{\alpha}$ and $d_{s}^{\beta}$, is denoted as \textit{$dp_{l,s}^{\alpha, \beta}$}, calculated in Equation \ref{eq_1}.
\begin{equation}
    \label{eq_1}
    dp_{l,s}^{\alpha, \beta} = d_{l}^{\alpha}  \odot d_{s}^{\beta} = \sum\limits_{i=1}^{128}{ f_{l,i}^{\alpha} \cdot f_{s,i}^{\beta}}
\end{equation}
Thus, \textit{$dp_{k}^{\alpha, \beta}$} is a dot-product of the $k^{th}$ descriptor of image $\Tilde{\alpha}$, with each descriptor of image $\Tilde{\beta}$, defined as
\[
dp_{k}^{\alpha, \beta}
=
\begin{bmatrix}
    dp_{k,1}^{\alpha, \beta}\\ 
    dp_{k,2}^{\alpha, \beta}\\
    \vdots \\ 
    dp_{k,n}^{\alpha, \beta} \\
\end{bmatrix}
=
d_{k}^{\alpha} \odot 
\begin{bmatrix}
    d_{1}^{\beta} \\ 
     d_{2}^{\beta} \\ 
    \vdots \\ 
    d_{n}^{\beta}
\end{bmatrix}
%\]
%\[
=
\begin{bmatrix}
    d_{k}^{\alpha}  \odot d_{1}^{\beta} \\ 
    d_{k}^{\alpha}  \odot d_{2}^{\beta} \\
    \vdots \\ 
    d_{k}^{\alpha}  \odot d_{n}^{\beta} \\
\end{bmatrix}
=
\begin{bmatrix}
    \sum\limits_{i=1}^{128}{ f_{k,i}^{\alpha} \cdot f_{1,i}^{\beta} }\\ 
    \sum\limits_{i=1}^{128}{ f_{k,i}^{\alpha} \cdot f_{2,i}^{\beta} }\\ 
    \vdots \\ 
    \sum\limits_{i=1}^{128}{ f_{k,i}^{\alpha} \cdot f_{n,i}^{\beta} }\\ 
\end{bmatrix}
.\]
Therefore, the dot-product of all descriptors of image $\Tilde{\alpha}$ and image $\Tilde{\beta}$ can be denoted as $dp^{\alpha, \beta}$, defined by
\[
dp^{\alpha, \beta}
=
\begin{bmatrix}
    dp_{1}^{\alpha, \beta}\\ 
    dp_{2}^{\alpha, \beta}\\
    \vdots \\ 
    dp_{m}^{\alpha, \beta} \\
\end{bmatrix}
%\]
%\[
=
\begin{bmatrix}
    dp_{1,1}^{\alpha, \beta} & dp_{2,1}^{\alpha, \beta} & dp_{3,1}^{\alpha, \beta} & \dots  & dp_{m,1}^{\alpha, \beta} \\
    dp_{1,2}^{\alpha, \beta} & dp_{2,2}^{\alpha, \beta} & dp_{3,2}^{\alpha, \beta} & \dots  & dp_{m,2}^{\alpha, \beta} \\
    \vdots & \vdots & \vdots & \ddots & \vdots \\
    dp_{1,n}^{\alpha, \beta} & dp_{2,n}^{\alpha, \beta} & dp_{3,n}^{\alpha, \beta} & \dots  & dp_{m,n}^{\alpha, \beta} 
\end{bmatrix}^{T}
.\]
In the SIFT matching algorithm the cosine inverse (arc-cosine), denoted by $ci$, of each dot-product operation is calculated. Similarly, $ci^{\alpha, \beta}$ is the arc-cosine of $dp^{\alpha, \beta}$, defined as
\[
ci^{\alpha, \beta}
=
\begin{bmatrix}
    ci_{1}^{\alpha, \beta}\\ 
    ci_{2}^{\alpha, \beta}\\
    \vdots \\ 
    ci_{m}^{\alpha, \beta} \\
\end{bmatrix}
=
\begin{bmatrix}
    ci_{1,1}^{\alpha, \beta} & ci_{2,1}^{\alpha, \beta} & ci_{3,1}^{\alpha, \beta} & \dots  & ci_{m,1}^{\alpha, \beta} \\
    ci_{1,2}^{\alpha, \beta} & ci_{2,2}^{\alpha, \beta} & ci_{3,2}^{\alpha, \beta} & \dots  & ci_{m,2}^{\alpha, \beta} \\
    \vdots & \vdots & \vdots & \ddots & \vdots \\
    ci_{1,n}^{\alpha, \beta} & ci_{2,n}^{\alpha, \beta} & ci_{3,n}^{\alpha, \beta} & \dots  & ci_{m,n}^{\alpha, \beta} 
\end{bmatrix}^{T}
.\]

\subsection{Software Approach of the SIFT Matching Algorithm} \label{sec:SW_ALOG}

The SIFT matching algorithm iterates through several steps to check if a match of a single descriptor of image $\Tilde{\alpha}$ corresponds with another descriptor in image $\Tilde{\beta}$. The efficient design of a matching algorithm depends largely on the platform in which implementation is to occur. In this section a software approach is detailed with a description of the resulting implementation of the SIFT matching algorithm based on our proposed angular distance measure between descriptors.

\begin{algorithm}[h]
 \caption{Software approach for SIFT descriptor matching.}
 \begin{algorithmic}[1]\label{alg_1}
 \renewcommand{\algorithmicrequire}{\textbf{Input:}}
 \renewcommand{\algorithmicensure}{\textbf{Output:}}
 \REQUIRE \textit{kth} descriptor of image $\alpha$ with size \textit{1 $\times$ 128}\\
Database descriptors of image $\beta$  with size \textit{m $\times$ 128}\\
 \ENSURE  Matched result for \textit{kth} descriptor with the database descriptors\\  
    \FOR {$j = 1$ to $m$}
        \FOR {$k = 1$ to $128$}
            \STATE $p[i][j] += A[i][k] \cdot  B[j][k]$;
        \ENDFOR 
       \STATE $[sort\_vals, index] = sort( arccos(p[i][j]) )$;
        \IF {$sort\_vals(1) < (threshold * sort\_vals(2))$} 
            \RETURN \textit{Match Found}
        \ELSE
            \RETURN \textit{No Match Found}
        \ENDIF
    \ENDFOR
 \end{algorithmic} 
\end{algorithm}

Algorithm \ref{alg_1} provides the computational software flow of the SIFT matching algorithm for the $k^{th}$ descriptor, $d_{k}^{\alpha}$, of image $\Tilde{\alpha}$ with the database descriptors of image $\Tilde{\beta}$, $d^{\beta}$, where image $\Tilde{\beta}$ has $n$ descriptors, represented as
\[
d^{\beta} = \{ d_{1}^{\beta}\ ,\ d_{2}^{\beta}\ ,\ \dots \dots\ ,\ d_{n}^{\beta} \}
.\]
The first step of the SIFT matching algorithm is to calculate the dot-product of the descriptor, $d_{k}^{\alpha}$, with each descriptor in the database according to Equation \ref{eq_1}. The result of the dot-product operation is a vector, \textit{$dp_{k}^{\alpha, \beta}$}, of \textit{n} elements, shown as
\[ 
dp_k^{\alpha, \beta} = [dp_{k,1}^{\alpha, \beta}\ , \  dp_{k,2}^{\alpha, \beta}\ ,\ \dots \dots\ ,\ dp_{k,n}^{\alpha, \beta} ]
.\]
The following step is to take the arc-cosine of each element in $dp_k^{\alpha, \beta}$, saving the result in memory or cache, presented mathematically as
\[ 
ci_k^{\alpha, \beta} = [ci_{k,1}^{\alpha, \beta}\ , \  ci_{k,2}^{\alpha, \beta}\ ,\ \dots \dots\ ,\ ci_{k,n}^{\alpha, \beta} ]
.\]
The resulting output array, $ci_k^{\alpha, \beta}$, is sorted in ascending order where the first and second minimums are calculated. 

David Lowe defined a threshold criteria \cite{DavidLowe1999}, typically $0.6$, to determine matching success. Matching success is determined by the match between the $k^{th}$ descriptor, $d_{k}^{\alpha}$, of image $\Tilde{\alpha}$ with the database descriptors, $d^{\beta}$, of image $\Tilde{\beta}$, according to Equation \ref{eq_matching}.

\begin{equation}
    \label{eq_matching}
    \begin{cases}
     \ minimum < ( 0.6\times second\_minimum) & \ \ \text{Match}\\ 
     \ otherwise  & \ \ \text{No Match}
    \end{cases}
\end{equation}

The calculations listed are repeated for each descriptor in image $\Tilde{\alpha}$ to determine the matching features in image $\Tilde{\beta}$. From Algorithm \ref{alg_1}, the SIFT matching algorithm requires an equally large number of calculations and memory resources; quickly showing large time dependency due to both calculation and memory access latency.
%%%%%%%%%%%%%%%%%%%%%%%%%%%%%%%%%%%%%%%%%%%%%%%%%%%%%%%%%

%%%%%%%%%%%%%%%%%%%%%%%%%%%%%%%%%%%%%%%%%%%%%%%%%%%%%%%%%
%%%%%%%%%%%%%%%%%%% Section %%%%%%%%%%%%%%%%%%%%%%%%%%%%%
%%%%%%%%%%%%%%%%%%%%%%%%%%%%%%%%%%%%%%%%%%%%%%%%%%%%%%%%%
\section{Proposed Optimization of Memory Bandwidth} \label{sec:comp_mem_BW}

In this section, we study the impact of the memory bandwidth on the overall performance of the matching process and explore an optimization scheme to fully utilize the computation core and the memory bandwidth.   

\subsection{Roofline Performance Model}

Image descriptors are streamed to the SIFT descriptor matching algorithm subsystem via an attached memory to the computing core. The total memory bandwidth plays a vital role in achieving maximum performance for a given system.
In order for the matching core to start processing, one descriptor for each image, $\Tilde{\alpha}$ and $\Tilde{\beta}$, should be ready at the input ports of the  matching core. We assume that the $k^{th}$ descriptor of image $\Tilde{\alpha}$ is always ready at the input port of the computational core. Since each descriptor is composed of 128 elements\footnote{Each element would be nominally composed of a 16-bit fixed point for the angular distance method.}, each data transfer between memory and the computational core is 256 bytes.

\begin{figure}[h]
	\centering
	\includegraphics[width=4.0in, keepaspectratio]{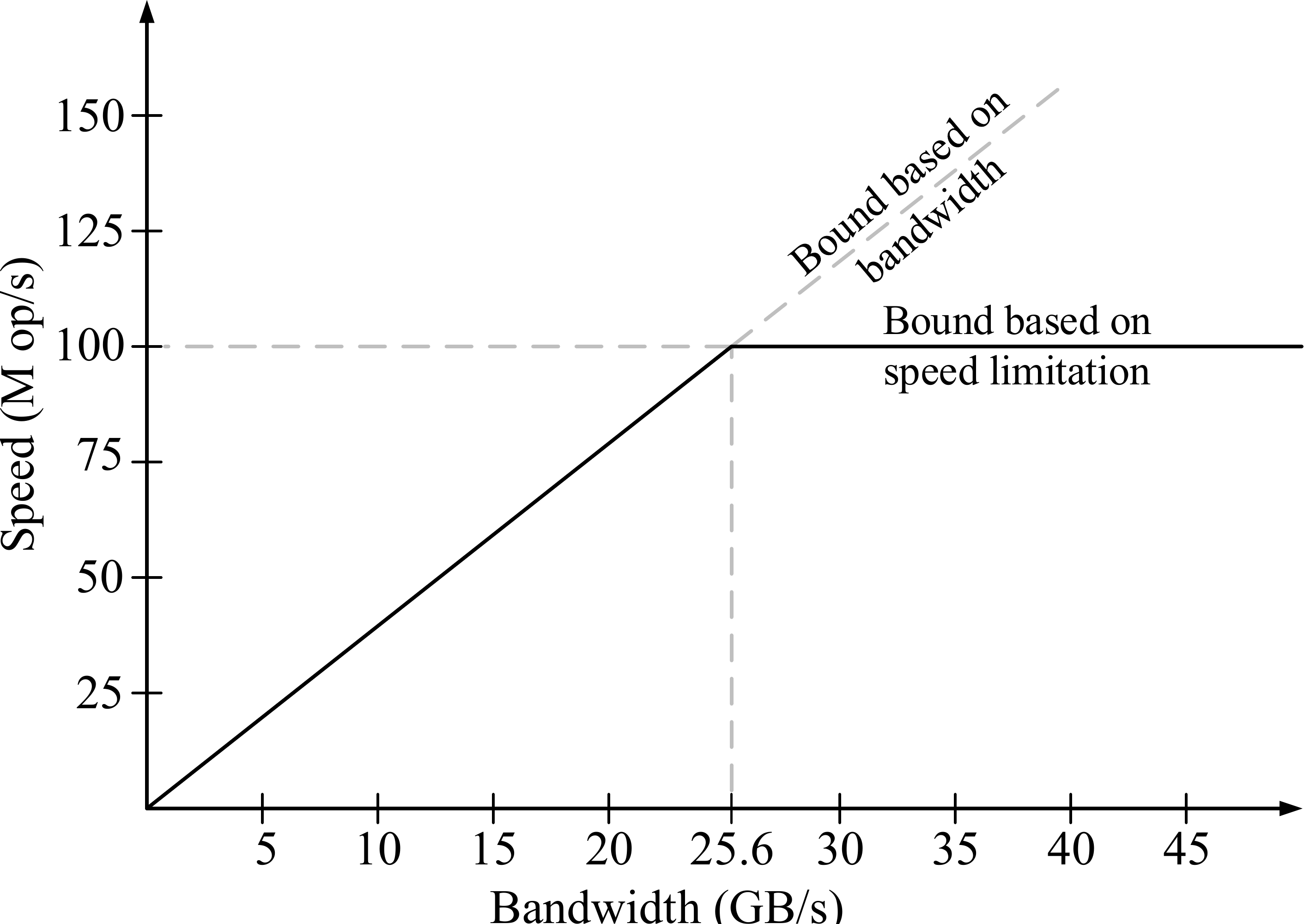}
	\caption{Performance and memory bandwidth effect.}
	\label{Fig_Roofline}
\end{figure}

In order to study the effect of the memory bandwidth in the overall performance of the system, let's assume that only one computational core exists in the system, that is pipelined and works at 100 MHz. To execute one operation, a full descriptor (256 bytes) should be ready at the input port of the computational core. Hence, the memory bandwidth take part in the system throughput. For example, if the memory bandwidth reaches 32 bytes per clock-cycle (3.2 GB/s), the computational core will wait for 8 clock-cycles to completely receives a single descriptor to start the process. This will achieve 12.5 Mega operation/second (M op/s). When the memory bandwidth increases to 6.4 GB/s, similarly, the performance increases to 24 M op/s. As long as the memory bandwidth increases, the performance increases. However, the maximum attainable throughput stops at its maximum peak when the memory bandwidth reaches 256 bytes per clock-cycle (25.6 GB/s) at the input port of the computation core. As the memory bandwidth increases above this limit, more data is present at the input port of the computation core but only 256 bytes are processed at a time. Figure \ref{Fig_Roofline} shows the effect of the memory bandwidth on the system performance. The speed of the computational core increases with increasing the memory bandwidth till it reaches the boundary of the peak performance.

%%%%%%%%%%%%%%%%%%%%%%%%   Figure#1     %%%%%%%%%%%%%%%%%%%%%%
%%%%%%%%%%%%%%%%%%%%%%%%%%%%%%%%%%%%%%%%%%%%%%%%%%%%%%%%%%%%%%

In this paper, we implemented the matching core on Zedboard \cite{zynqboardmanual}. The platform includes two DDR3 memory components. The multi-protocol DDR controller is configured for 32-bit wide accesses to a 512 MB address space. For 32-bit data width access of the DDR memory, 64 bits (8 bytes) are accessed in one clock-cycle. This limits the performance of the matching core to 100/32 M op/s for 100 MHz running clock, where the core waits for 32 clock-cycles to receive a complete descriptor to start its operation. However, by optimizing the memory access, we can achieve the peak performance of the platform, 100 M op/s, as illustrated in section \ref{memory_optimization}.

\subsection{Memory Access Optimization} \label{memory_optimization}

Due to the maximum memory bandwidth limitations presented by the hardware platform, our goal is to increase throughput by executing one dot-product operation every clock-cycle. To assure one dot-product operation can be computed every clock-cycle, a new descriptor must be valid every clock-cycle. In order to alleviate the memory bandwidth bottleneck, an internal memory (cache) is used for storing 32 descriptors. Since the time to calculate 32 dot-product operations is 32 clock-cycles (one operation per clock-cycle), within that time period, one complete descriptor can be fetched from external memory. The newly fetched descriptor will execute a dot-product operation with each descriptor in the internal cache (32 descriptors stored in internal cache). The result of the fetching optimization operations allows calculating 32 dot-product operations in 32 clock-cycles. While executing the 32 dot-products of the current descriptor, a new descriptor is received and the process is repeated until the entirety of descriptors of image $\Tilde{\beta}$ is completed. 

Therefore, in order to alleviate the memory bandwidth restriction, the descriptors of image $\Tilde{\alpha}$ are divided into blocks of 32 descriptors each. Each block is passed to an internal cache and the dot-product operation is executed with one block and the entirety of descriptors of image $\Tilde{\beta}$. Architecturally, two first-in first-out (FIFO) buffers are used to store the descriptors from external memory as a linear cache, with a total fetch time of 1024 clock-cycles per block\footnote{1024 clock-cycles comes from the previous 32 clock-cycles to fetch a single descriptor by the number of descriptors per block, $32\times 32 = 1024$.}. The result from the block latency is that the highest throughput can be achieved when image $\Tilde{\beta}$ has more than 32 descriptors. To further alleviate computation time and memory requirements for the SIFT descriptors matching, the architecture must be fully compatible with the platform. In our approach, the matching architecture is implemented such that full utilization of the core is achieved.
%%%%%%%%%%%%%%%%%%%%%%%%%%%%%%%%%%%%%%%%%%%%%%%%%%%%%%%%%

%%%%%%%%%%%%%%%%%%%%%%%%%%%%%%%%%%%%%%%%%%%%%%%%%%%%%%%%%
%%%%%%%%%%%%%%%%%%% Section %%%%%%%%%%%%%%%%%%%%%%%%%%%%%
%%%%%%%%%%%%%%%%%%%%%%%%%%%%%%%%%%%%%%%%%%%%%%%%%%%%%%%%%
\section{Proposed Matching Algorithm Architecture on Hardware} \label{sec:proposed_HW}

FPGA is generally utilized for accelerating computational processes by increasing concurrent operations. It further increases the overall throughput of the system by pipelining and overlapping the instructions. The goal of this work is to accelerate the SIFT descriptors matching on FPGA and efficiently handling memory bandwidth limitations which is often seen in software implementation, as explained in Section~\ref{sec:comp_mem_BW}.

In our proposed architecture, the descriptors of image $\Tilde{\alpha}$ and image $\Tilde{\beta}$ are streamed from external memory into an internal FIFO. Each descriptor is composed of 128 elements and its location, $(x,y)$, in the image. Although each element should be represented as a double-precision floating-point to increase the accuracy, such floating-point adder circuits \cite{daoud2015survey} is more complicated and consumes more resources. Therefore, for further optimization, each element of the descriptor is represented as a 16-bit fixed-point value and a total of 32-bits for its location, leading to an individual descriptor size of 2080 bits. 

The proposed SIFT matching architecture \cite{daoud2018sift} consists of four main sub-cores and two internal caches to alleviate memory bottlenecks; all of which are fully pipelined and implemented onto FPGA:
\begin{itemize}
    \item Dot\_Product.
    \item Cosine\_Inverse.
    \item Minimum Search (MIN\_FIND).
    \item Match\_Check.
    \item Descriptor Cache (DES\_MEM).
    \item Minimum(s) Cache (MIN\_MEM). 
\end{itemize}
For making use of high-level synthesis design \cite{daoud2014survey}, Xilinx System Generator \textsuperscript{\textregistered} is used for designing and implementing the \textit{Dot\_Prod} and the \textit{Cosine\_Inverse} blocks as IP cores.

\begin{figure}[h!]
	\centering
	\includegraphics[width=\columnwidth, keepaspectratio]{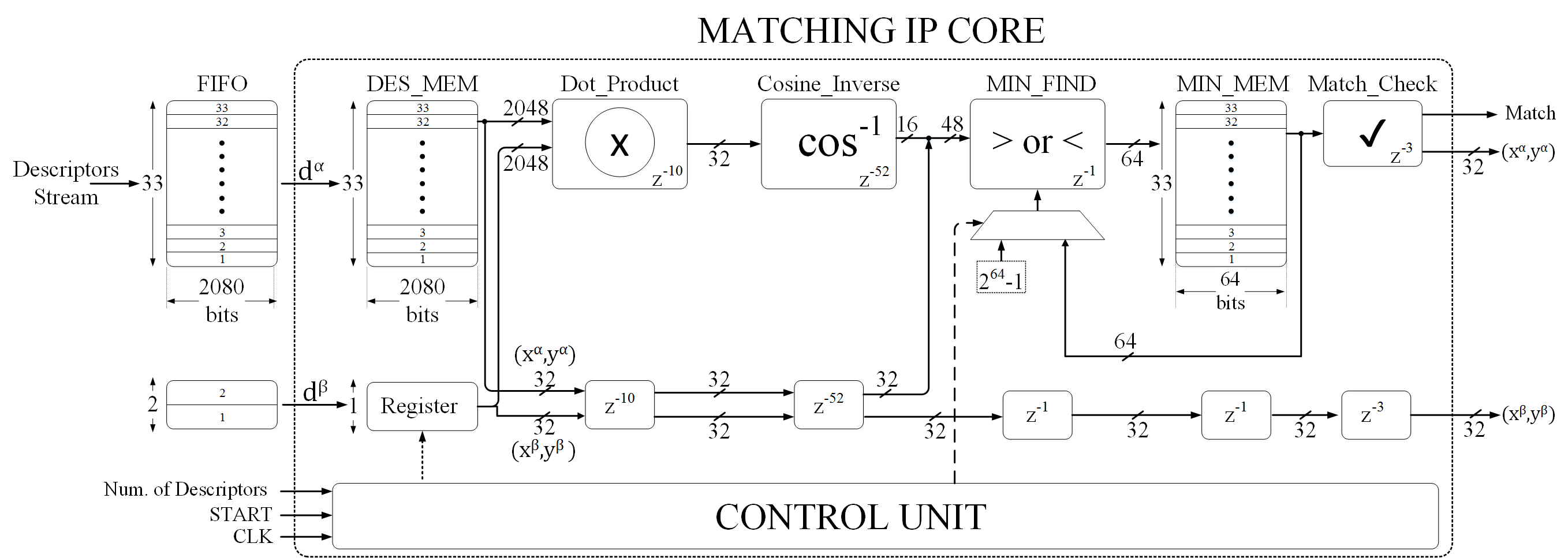}
	\caption{Matching core architecture for the proposed SIFT descriptors matching detailing the sub-modules and caches necessary for operation as controlled by the control unit. $Z^{-n}$ is a shift register with n pipeline stages.}
	\label{Fig_CompleteBlockDiagram}
\end{figure}

Figure \ref{Fig_CompleteBlockDiagram} describes a block diagram of our proposed SIFT descriptor matching accelerator core. In this architecture, the descriptors are streamed from the memory to the matching core. The internal caches (buffers) are used for keeping the descriptors to alleviate the memory bottleneck and fully utilize the matching core. This will allow a complete descriptor available at every clock-cycle, where the external memory bandwidth is optimized for increasing the throughput as explained in Section \ref{memory_optimization}.

%%%%%%%%%%%%%%%%%%%%%%%%   Figure#2     %%%%%%%%%%%%%%%%%%%%%%
%%%%%%%%%%%%%%%%%%%%%%%%%%%%%%%%%%%%%%%%%%%%%%%%%%%%%%%%%%%%%%

Since the descriptor size is 260 bytes (including the coordinates) and the memory bandwidth is 8 bytes per clock, it takes 33 clock-cycles to fetch a complete descriptor. In order to match the fetching time, 33 descriptors of image $\Tilde{\alpha}$ were saved in a separate descriptors cache (DES\_MEM), where each descriptor of image $\Tilde{\beta}$ falls through in 33 clock-cycles and saved into Register.

The outputs of the DES\_MEM and the Register are split into descriptors that are handed to the Dot\_Product core and their coordinates that are passed through shift register with a corresponding number of the pipeline stages. The output of the Dot\_Product is passed to a Cosine\_Inverse core of 52 pipeline stages approximating the resulted output to 16-bits. 

To calculate the minimum and second minimum on the fly, the previous minimum and second minimum are retrieved from the minimum(s) cache (MIN\_MEM)\footnote{Initially, the MIN\_MEM is empty; the minimum and second minimum values are delivered and temporarily considered as the maximum values ($0\times FFFF$).} and passed to the minimum-search core (MIN\_FIND) along with the current calculated cosine-inverse.

For each new descriptor block of image $\Tilde{\alpha}$, MIN\_MEM should be flushed. Therefore, a multiplexer is used to pass a constant value of ($0\times FFFF$) when a new block of descriptor is delivered at DES\_MEM. Otherwise, it passes the current value(s) in MIN\_MEM. This is controlled by the Control Unit, seen in Figure \ref{Fig_CompleteBlockDiagram}.

The Control Unit present in the proposed SIFT matching architecture allows several operations to run concurrently by using a scheduling method; increasing overall throughput of the system. An example of scheduled concurrent operation is when the execution of a dot-product operation occurs on the final descriptor of image $\Tilde{\beta}$, the DES\_MEM is simultaneously filled with the subsequent descriptor of image $\Tilde{\alpha}$ such that the following descriptor of image $\Tilde{\beta}$ will already have a new reference descriptor block. The Control Unit is aware of the total number of descriptors and the processing time of each core, which make it able to handle the control signals to receive new descriptors, enable the internal cores, and control the internal caches.

\subsection{Dot\_Product Core}
\begin{figure}[!htb]
	\centering
	\includegraphics[width=3.0in,keepaspectratio]{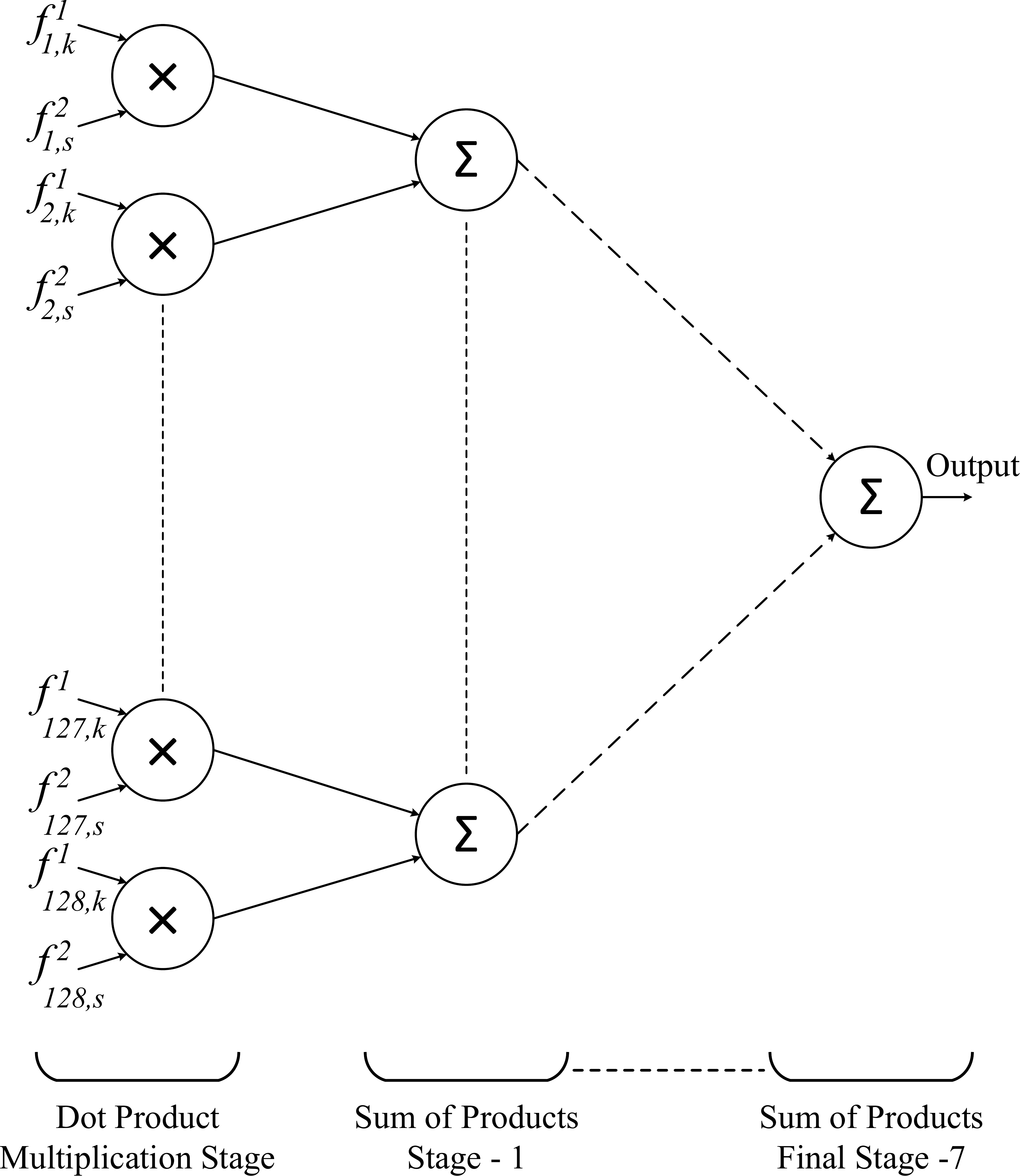}
	\caption{Dot product core.}
	\label{Fig_DotProductTree}
\end{figure}

The SIFT matching algorithm begins by calculating the product of each element of the $k^{th}$ descriptor of image $\Tilde{\alpha}$ with the corresponding element of image $\Tilde{\beta}$. Since the descriptor is comprised of 128 elements, 128 multiplications are required to calculate them in parallel. The output from each prior multiplication is sequentially added to obtain the resulting dot-product by the Dot\_Product core which composed of 128 multiplication cores and seven levels\footnote{Seven levels of adders are required since $log_{2}{128}=7$, meaning for each tree we can compute a segment of the 128 elements.} of tree adders. The multipliers necessary for the Dot\_Product core were implemented into the FPGA's digital signal processing (DSP) slices with 3 pipeline stages. The seven levels of adders necessary for sequential addition are equivalent to 7 pipeline stages. Figure \ref{Fig_DotProductTree} shows the internal architecture of the Dot\_Product core with a total of 10 pipeline stages.

%%%%%%%%%%%%%%%%%%%%%%%%   Figure#3     %%%%%%%%%%%%%%%%%%%%%%
%%%%%%%%%%%%%%%%%%%%%%%%%%%%%%%%%%%%%%%%%%%%%%%%%%%%%%%%%%%%%%

%%%%%%%%%%%%%%%%%%%%%%%%   Figure #4     %%%%%%%%%%%%%%%%%%%%%%

\begin{figure}[!htb]
	\centering
	\includegraphics[width=4.5in,keepaspectratio]{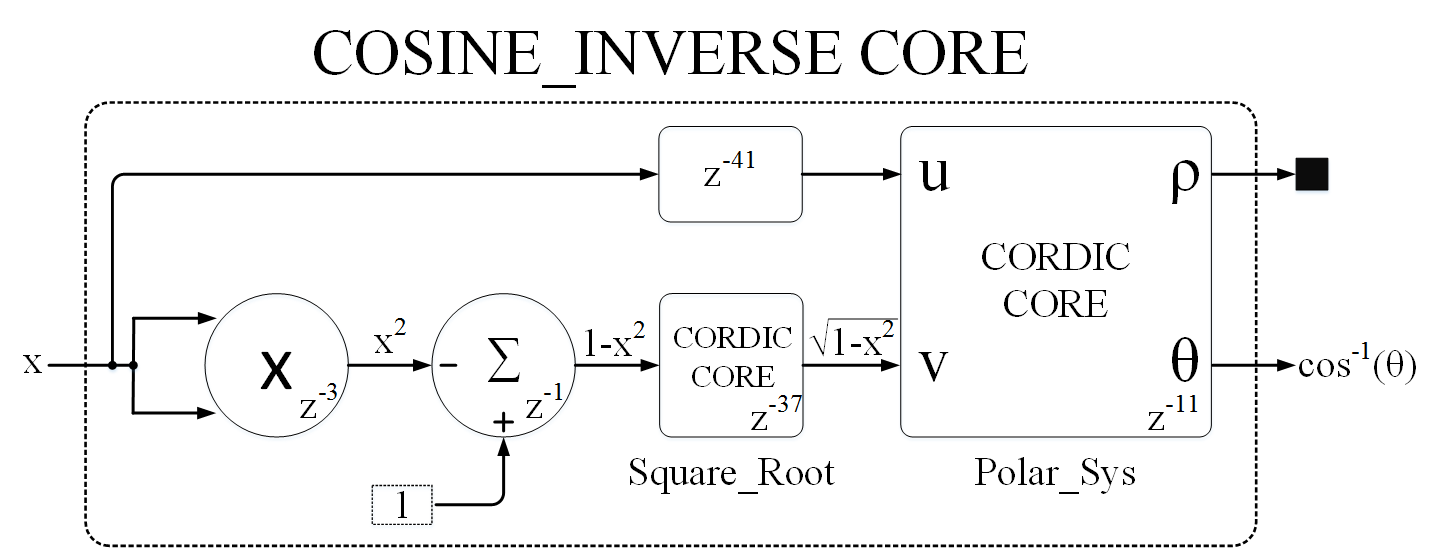}
	\caption{Internal depiction of the ''Cosine\_Inverse'' core, including square root calculation and the polar coordinate translation module.  $Z^{-n}$ is a shift register with n pipeline stages.}
	\label{Fig_CosineInverse}
\end{figure}
%%%%%%%%%%%%%%%%%%%%%%%%%%%%%%%%%%%%%%%%%%%%%%%%%%%%%%%%%%%%%%

\subsection{Cosine Inverse Core}

In order to implement the cosine-inverse in hardware, a coordinate rotation digital computer (CORDIC) core provided by Xilinx using System Generator for DSP \cite{systemgenerator} is used. The Cosine\_Inverse core is shown in Figure \ref{Fig_CosineInverse}. It is composed of two CORDIC cores: one to calculate the square root of an input and another to find the polar coordinates of a feature, labeled in Figure \ref{Fig_CosineInverse} as Square\_Root and Polar\_Sys, respectively. Internal to the Square\_Root and Polar\_Sys core are 37 and 11 pipeline stages, respectively. The calculation of $1-x^{2}$ as an input to the Square\_Root core is completed with 4 pipeline stages, with a total of 52 pipeline stages for the Cosine\_Inverse core alone.

\subsubsection{Polar\_Sys Core}

The Polar\_Sys core within the Cosine\_Inverse core has two inputs, \textit{u}, and \textit{v} of a Cartesian system, and two outputs, magnitude, $\rho$, and angle, $\theta$. The relation between these Cartesian system, $(u, v)$ and the polar system, ($\rho$, $\theta$) is simply described for a right triangle with hypotenuse $\rho$ and sides $u$ and $v$ as $\rho = \sqrt{u^2 + v^2}$ and $\theta = tan^{-1}(v/u)$. To obtain the arc-cosine, only the calculation of $\theta$ is necessary, thus only computing $\theta=cos^{-1}x$, where $x$ is an input into the Cosine\_Inverse needs calculation. The two inputs of the Polar\_Sys core, $(u, v)$ are thus obtained as $u = x$ and $v= \sqrt{1 - x^{2}}$.

\subsection{Minimum Search (MIN\_FIND) Core}

After calculating the arc-cosine for each descriptor in the database, two minimum values are determined, serving to highlight the database descriptors that are potential candidates for similarity within the image's descriptor under consideration. In software approach to descriptor matching, the output values are stored into a memory then a sorting algorithm is applied to find the minimum and second minimum values. In hardware approach to descriptor matching, a typical sorting algorithm is resource-inefficient due to memory needs thus, in our design, the MIN\_FIND core is designed to find both the minimum and second minimum values on the fly. This is done by retrieving the previous minimum and second minimum values from (MIN\_MEM) and compared with the current calculated cosine-inverse. The pseudo-code representing the hardware operation of the  MIN\_FIND core is shown in Algorithm \ref{algcomp}. 

\begin{algorithm}[ht]
 \caption{Comparison scheme for calculating minimum and second minimum values to highlight database descriptors as potential candidates for similarity with the image descriptor.}\label{algcomp}
 \begin{algorithmic}[1]
 \renewcommand{\algorithmicrequire}{\textbf{Input:}}
 \renewcommand{\algorithmicensure}{\textbf{Output:}}
 \REQUIRE Output of the cosine inverse (curr\_val), recent minimum value from the memory (prev\_min), and recent second minimum value (prev\_sec\_min).

 \ENSURE Updated minimum value (min) and second minimum value (sec\_min).

  \IF {$curr\_val < prev\_min$} 
    \STATE $min = curr\_val $;\\
    \STATE $sec\_min = prev\_min$;\\ 
  \ELSIF{{$curr\_val < prev\_sec\_min$} }
    \STATE $min = prev\_min$;\\
    \STATE $sec\_min = curr\_val$;\\ 
  \ELSE
    \STATE $min = prev\_min$;\\
    \STATE $sec\_min = prev\_sec\_min$;\\
  \ENDIF

 \RETURN \textit{min, sec\_min}
 \end{algorithmic} 
\end{algorithm}

\subsection{Match\_Check Core}

\begin{figure}[h!]
	\centering
	\includegraphics[width=3in,keepaspectratio]{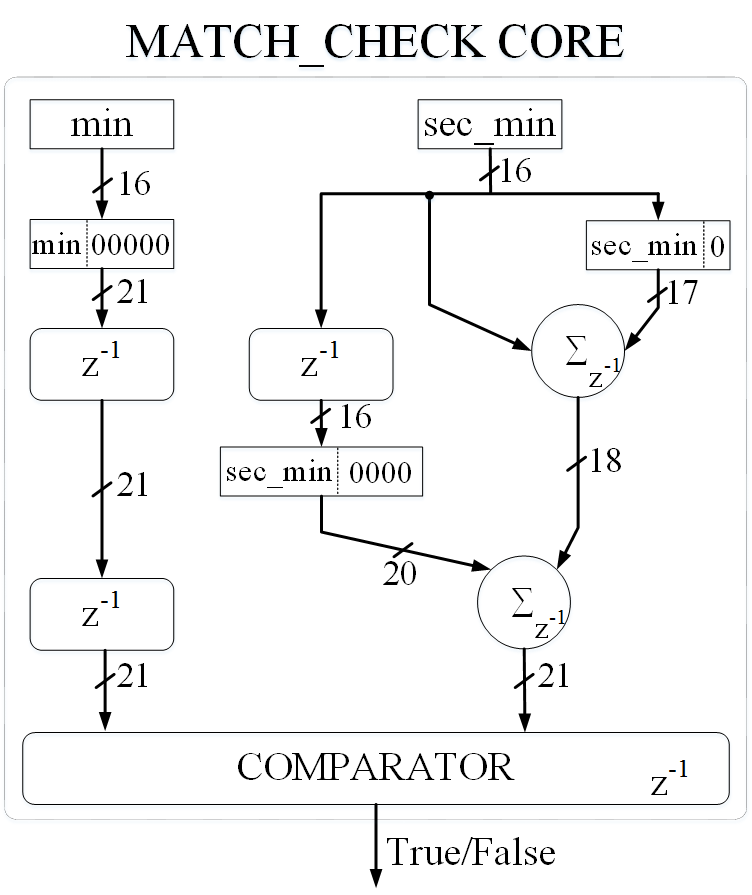}
	\caption{Matching Check core.}
	\label{Fig_Comparator}
\end{figure}

The final step of the SIFT matching algorithm is to check if an actual match occurs by passing the minimum and second minimum values to the Match\_Check core. The Match\_Check core then applies Equation \ref{eq_matching} to check matching between the descriptor of image $\Tilde{\alpha}$ with another descriptor within image $\Tilde{\beta}$. The hardware design of the Match\_Check core consists of 3 pipeline stages \textit{without} the need for any multiplier. The multiplication procedure of the second minimum with 0.6 is hidden in an addition process since $(0.6)_{d}=($0.10011$)_{b}$ can be used as a constant value. Therefore, $\textit{minimum} \times (100000)_{b}$ is compared with $\textit{second minimum} \times (10011)_{b}$ to check matching between the two descriptors, previously mentioned in Algorithm \ref{alg_1}. Figure~\ref{Fig_Comparator} shows a three pipelined stages of adder circuits to implement Equation \ref{eq_matching}, where the multiplication of a number with the constant value $($10011$)_{b}$ is done by adding the number to itself after it is shifted to the left one time and the result is added to the same number after it is shifted to the left four times. The shifting process was done by appending the right side of the number with extra zero-bits. 

%%%%%%%%%%%%%%%%%%%%%%%%   Figure #5    %%%%%%%%%%%%%%%%%%%%%%
%%%%%%%%%%%%%%%%%%%%%%%%%%%%%%%%%%%%%%%%%%%%%%%%%%%%%%%%%%%%%%

%%%%%%%%%%%%%%%%%%%%%%%%%%%%%%%%%%%%%%%%%%%%%%%%%%%%%%%%%
%%%%%%%%%%%%%%%%%%% Section %%%%%%%%%%%%%%%%%%%%%%%%%%%%%
%%%%%%%%%%%%%%%%%%%%%%%%%%%%%%%%%%%%%%%%%%%%%%%%%%%%%%%%%
\section{Experimental Results and Evaluation} \label{sec:exp_eval}

In order to evaluate the proposed hardware architecture of the the SIFT descriptors matching, we used a Xilinx\textsuperscript{\textregistered} Zynq-7000-based Zedboard. The Zedboard has a programmable logic and two ARM Cortex-A9 co-processors. The SIFT hardware matching algorithm core was implemented into the Zedboard's programmable logic while a single ARM Cortex-A9 processor was only used for simulation within the Xilinx\textsuperscript{\textregistered} Software Development Kit.

\subsection{Experimental Setup}
In order for the SIFT matching core to start processing, descriptors  $d^{\alpha}$ and  $d^{\beta}$ of both images $\Tilde{\alpha}$ and  $\Tilde{\beta}$ should be ready at the input ports of the matching core. The descriptors for both images were initially stored onto a SD card used within the Zedboard, whose contents were then loaded into the external DRAM by the on-board processing system (PS). To check the matches between two images, the PS initializes and facilitates direct memory access (DMA) transfer of the provided descriptors from the DRAM to the descriptor buffers.

The PS then initiates the matching process with the matching core over advanced extensible interface (AXI) whereupon the matching core is provided with the number of descriptor blocks, number of descriptors per block, and a start signal. The Xilinx\textsuperscript{\textregistered} EDA design Suite was used to synthesize and implement the design, including the matching core, DMA, and FIFO buffers. The Xilinx\textsuperscript{\textregistered} Software Development Kit was used to read descriptors for both images from SD card into memory, and pass them to the fabric buffers/descriptor(s) caches. The fabric clock of the Zedboard has a range of 100 kHz to 250 MHz, however the AXI DMA has a maximum frequency of 150 MHz or 120 MHz for AXI4 and AXI4-Lite, respectively \cite{axidmamanual}. Due to the frequency limitations of the Zedboard's systems, all experiments were run at a nominal frequency of 100 MHz. 

\subsection{Experimental Results}
The SIFT matching algorithm was fully synthesized and implemented onto the Zedboard's fabric with a 135 MHz maximum frequency with a normal clock-speed of 100 MHz\footnote{Used in this context due to limitation of the DMA core and AXI4-Lite provided by the Vivado toolset.}. The implemented SIFT matching core has only one computational element of the ``Dot\_Product'', ``Cosine\_Inverse'', ``MIN\_FIND'', and ``Match\_Check '' cores. The matching core includes a ''Control\_Unit'', additional internal memories for Descriptors and Minimum(s) Caches, and other registers for pipelining and synchronizing the data flow of the algorithm. Table \ref{table_exp} summarizes the overall resource utilization for individual components, with the ``Others'' category collecting registers and multiplexers used for synchronizing descriptors and control signals.

%%%%%%%%%%%%%%%%%%%%%%%%   Table #1    %%%%%%%%%%%%%%%%%%%%%%
\begin{table}[ht]
\caption{Our proposed SIFT matching algorithm architecture utilization report for Zedboard FPGA implementation.}
\label{table_exp}
\centering
\begin{tabular}{|l|c|c|c|c|}
\hline
\rule{0pt}{9pt}
%\hline
\bf{Core} & \bf{LUT} & \bf{FF}  & \bf{DSP}  & \bf{BRAM} \\ 
%\hline

\hline
\rule{0pt}{9pt}
\bf{Dot-Product \& Cosine Inverse}  & 3382  & 3557 & 132 & 0\\
\rule{0pt}{9pt}
\bf{Minimum Search} & 82 & 66 & 0 & 0\\
\rule{0pt}{9pt}
\bf{Matching Check} & 42 & 283 & 0 & 0\\
\rule{0pt}{9pt}
\bf{Control Unit} & 92 & 2132 & 0 & 0\\
\rule{0pt}{9pt}
\bf{Descriptor Cache} & 0 & 0 & 0 & 29\\
\rule{0pt}{9pt}
\bf{Minimum(s) Cache} & 0  & 0 & 0 & 1\\
\rule{0pt}{9pt}
\bf{Others}  & 112 & 327 & 0 & 0\\

\hline
\hline
\rule{0pt}{9pt}
Total & 3710 & 6365 & 132 & 30\\
\hline
\end{tabular}
\end{table}
%%%%%%%%%%%%%%%%%%%%%%%%%%%%%%%%%%%%%%%%%%%%%%%%%%%%%%%%%%%%%%

\begin{figure*}[ht]
	\centering
	\subfloat[\textit{Image\_1} matching points (579 descriptors).]{\includegraphics[width=3in]{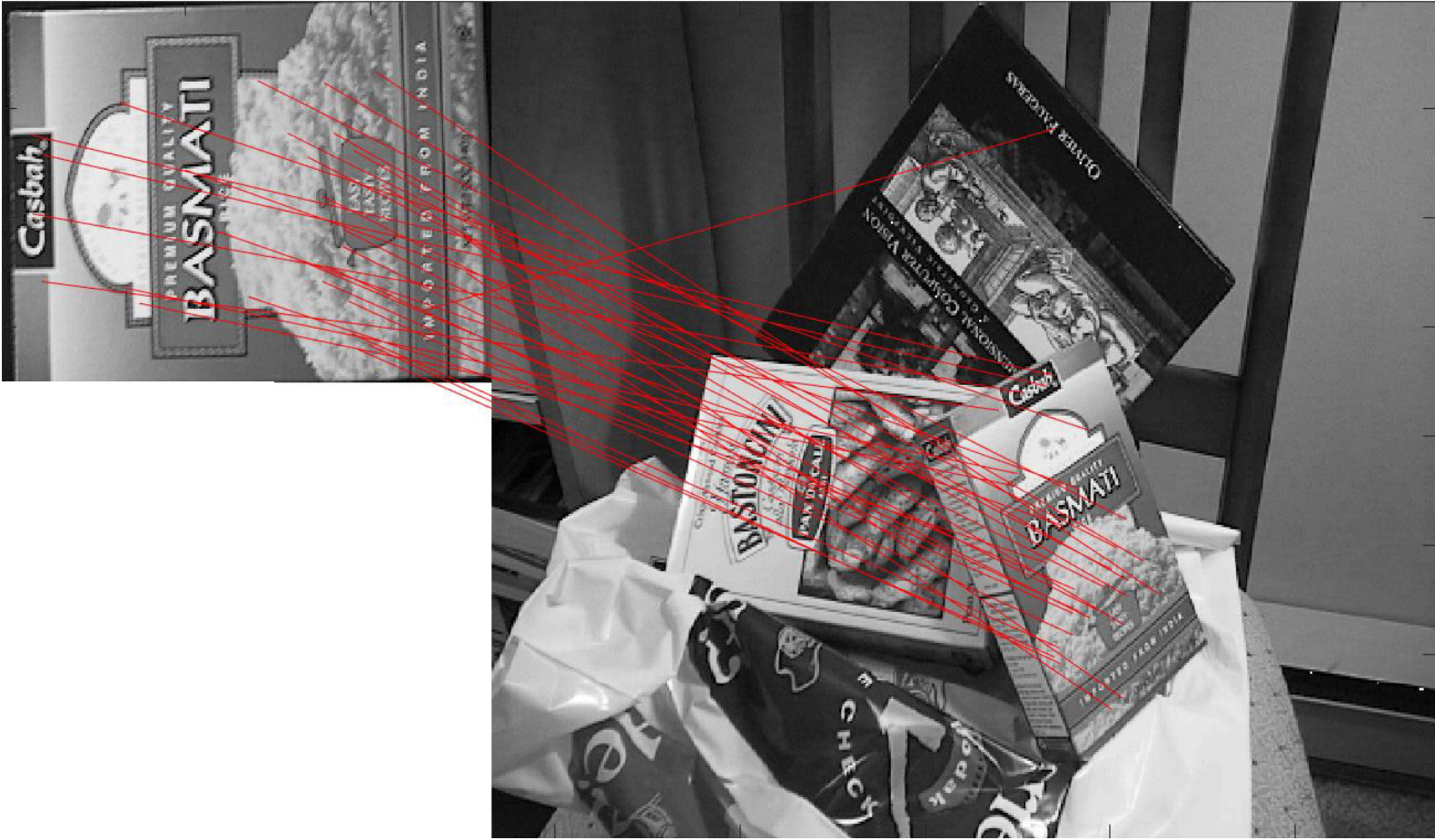}%
		\label{Fig_image1}}
	\hfil
	\subfloat[\textit{Image\_2} matching points (638 descriptors).]{\includegraphics[width=3in]{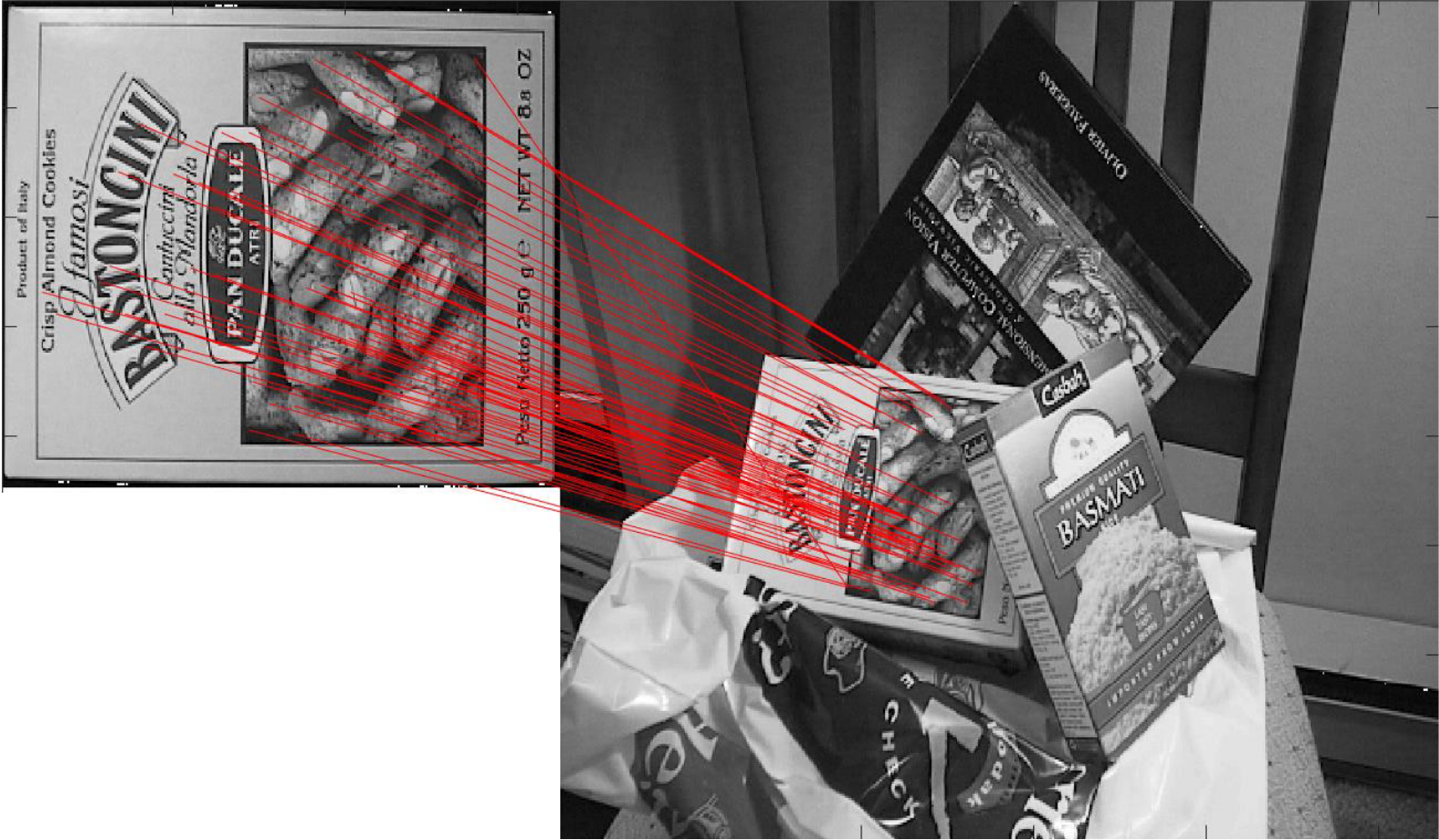}%
		\label{Fig_image2}}
	\hfil
	\subfloat[\textit{Image\_3} matching points (882 descriptors).]{\includegraphics[width=3in]{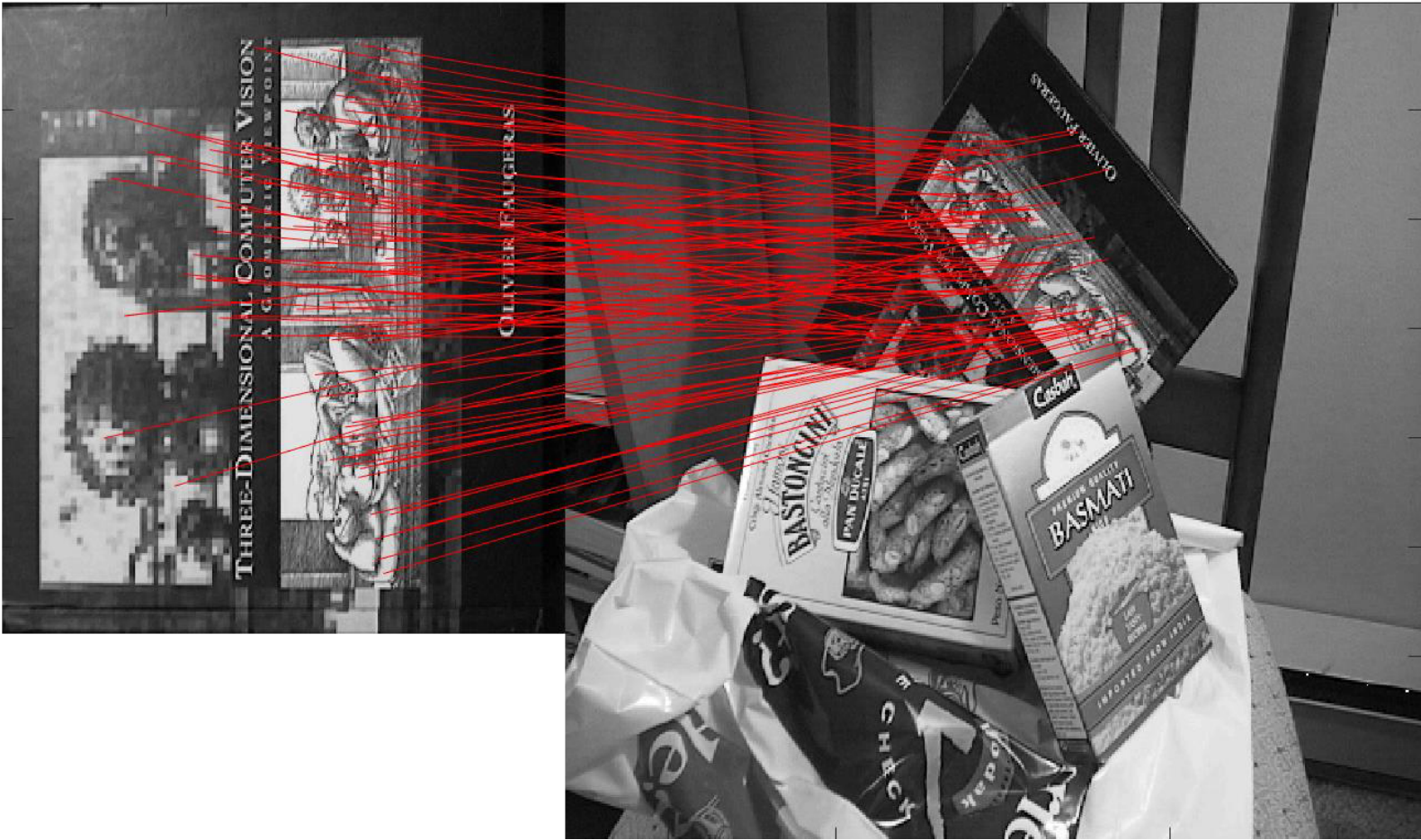}%
		\label{Fig_image3}}
	\hfil
	\subfloat[\textit{Image\_4} matching points (1021 descriptors).]{\includegraphics[width=3in]{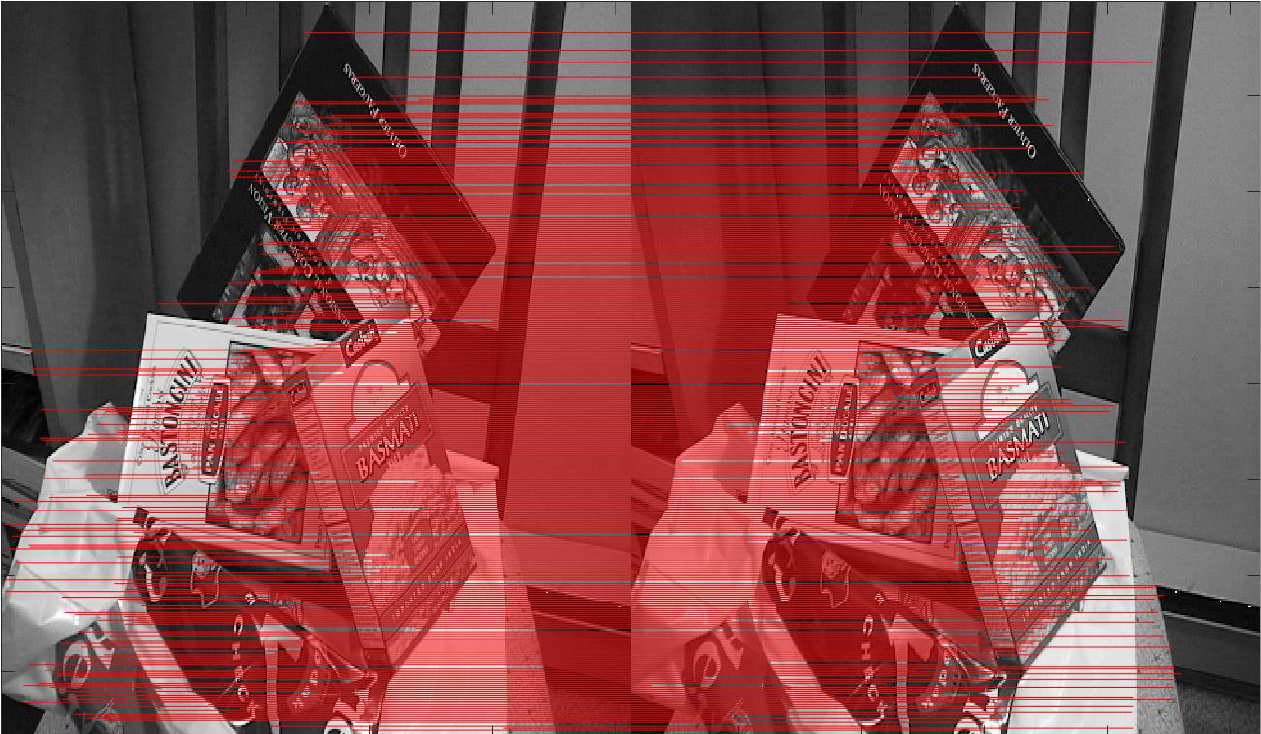}%
		\label{Fig_image4}}
	\caption{Matching points for selected images with different number of descriptors}\label{images}
\end{figure*}

In order to evaluate our SIFT matching core based on cosine angle distance approach, several experiments were conducted. The experiments were run on four different images, \textit{image\_1}, \textit{image\_2}, \textit{image\_3} and \textit{image\_4} where each image has 579, 538, 882, and 1021 descriptors, respectively. We chose \textit{image\_4} as the database image which the other three images were checked against for potential matches. To check the correctness of the matching points, \textit{image\_4} was used for testing the self-matching ability of the SIFT matching core as developed. Figure \ref{images} shows the matching points between the selected images with the database.

%%%%%%%%%%%%%%%%%%%%%%%%   Figure #6    %%%%%%%%%%%%%%%%%%%%%%
%%%%%%%%%%%%%%%%%%%%%%%%%%%%%%%%%%%%%%%%%%%%%%%%%%%%%%%%%%%%%%

For comparison of the descriptor matching time using a traditional software approach, the SIFT matching algorithm was executed on a 64-bit Intel\textsuperscript{\textregistered} Core 2 Duo CPU running at 3.16 GHz using MATLAB\textsuperscript{\textregistered} 2017a, following the original design described by Section \ref{sec:SW_ALOG}. By using our proposed hardware SIFT matching core, it took 6.08, 6.75, 9.11, and 10.46 milliseconds for images with 579, 638, 882, and 1021 descriptors, respectively, whereas the software approach took 71.6, 77.4, 105.6 and 163.9 milliseconds, respectively, for the same set of images. The differences in time taken for both the software and our proposed hardware approach can be summarized that the software approach takes a quadratic increase in computational time with an increasing number of descriptors, compared with a linear increase for our proposed algorithm. 

Our SIFT matching core accelerated the computational time of the selected images by (11.5 $\sim$ 15.7)$\times$ for (579 $\sim$ 1021) descriptors. The double-floating point operations within the SIFT matching algorithm in software were approximated to 16-bit fixed-point operations in hardware with 98\% of matched-descriptors detected with decreasing error for increasing fixed-point resolution. For further analysis, we compared our proposed hardware architecture of the SIFT matching algorithm with a similar approach accelerated on a graphics processing unit (GPU) in \cite{fassold2015gpu}. The Authors in \cite{fassold2015gpu} used both NVIDIA's Tesla K20 GPU and their local Intel Xeon\textsuperscript{\textregistered} 2.7 Ghz Quad-Core CPU with a set of 2,800 descriptors for both the test image and database set. The authors noted that their matching algorithm takes 13 milliseconds on GPU and 80 milliseconds on CPU. Our hardware approach takes 10.46 milliseconds for 1021 descriptors using only one computation core, compared with the 2,496 cores present in the K20. It is noteworthy that by increasing the computational cores with fixed memory bandwidth, the throughput of our system is not affected due to memory bandwidth limitation. In this presented work, we achieved the maximum performance by hitting the roofline performance model, i.e. achieving high throughput with the maximum allowed memory bandwidth and one computational core.

In addition, it is important to compare our proposed cosine angle distance approach for descriptor matching with other implementations in the literature. There are several matching techniques based on different approaches such as calculating the Chi-square distances \cite{comparisonPaper2017}, Sum of Absolute Differences (SAD) \cite{Vourvoulakis}, and calculating Hamming distances \cite{KAPELA}. 

The implementation in \cite{comparisonPaper2017} used 64 cores to calculate distance metrics in parallel, requiring many multiplications and divisions. The comparison of resources to our proposed implementation shows a  reduction of 91\% of LUTs, 79\% of BRAM. Apart from hardware resource utilization, our proposed SIFT matching core has the capability to check match-points within 9.28 milliseconds while \cite{comparisonPaper2017} takes 10 milliseconds for 900 descriptors, due to pipelining and maximization of memory bandwidth within the computation core.

The implementation in \cite{Vourvoulakis} used 16 SAD (Sum of Absolute Differences) calculators to calculate the absolute difference between descriptors. The 16 SAD values are passed through 4 levels of comparators to obtain the minimum SAD value as a potential candidate among 16 descriptors. They used 2 separate RAMs to store the intermediate descriptors.The comparison of resources compared to our proposed implementation results in a reduction of about 92\% of LUTs, 98\% of BRAM and 75\% DSP area. 

The implementation in \cite{KAPELA} used a variable number of Hamming distance calculators in the matching core. For implementation using 2 cores of Hamming distance a considerable saving in the FPGA resources were obtained. Compared to 2 cores to calculate Hamming distance, our implementation of the matching core using the cosine angle distance  approach saved 37\% of LUTs and 89\% of BRAM. As the number of Hamming distance calculation cores increases, the number of resources utilization increases. Table \ref{table_comparison2} shows the utilized resources of our proposed SIFT matching core using cosine angle distance  versus other matching methods such as Chi-square distance \cite{comparisonPaper2017}, SAD calculators \cite{Vourvoulakis} and Hamming distance \cite{KAPELA}.

%%%%%%%%%%%%%%%%%%%%%%%%   Table #2    %%%%%%%%%%%%%%%%%%%%%%
\begin{table}[h!]
\caption{Utilized resources of our architecture core vs other implementation in the literature.}
\label{table_comparison2}
\centering
\begin{tabular}{| l | l | l | l | l |}
\hline
\thead{\textbf{Parameter}} & 
\thead{\textbf{\cite{comparisonPaper2017}}} &
\thead{\textbf{\cite{Vourvoulakis}}} &
\thead{\textbf{\cite{KAPELA}}} &
\thead{\textbf{Proposed}}
\\ \hline
FPGA used             & Virtex6           &  Cyclone IV         & Zynq-7000      & Zynq-7000  \\ \hline
Image Size            & 512 $\times$ 384  &  640 $\times$ 480   &                & 512 $\times$ 384  \\ \hline
Type of descriptors   & SIFT              &  SIFT               & FREAK          & SIFT            \\ \hline
\# of descriptors     & 900               &  --                 & --             & 882             \\ \hline
LUTs (\% saved)       & 42662 (91\%)      &  51068 (92\%)       & 5967 (37\%)    & 3710            \\ \hline
DSP (\% saved)        & 104  (-27\%)      &  528  (75\%)        & --             & 132             \\ \hline
BRAM (\% saved)       & 142  (79\%)       &  213 (85\%) \tablefootnote{The authors reported 1697 kbits of memory, which is equivalent to 213 BRAM in the best case of memory utilization.}        & 294 (75\%)     & 30              \\ \hline
Clock Frequency (MHz) & 172               &  100                & 100            & 100             \\ \hline    
\end{tabular}
\end{table}
%%%%%%%%%%%%%%%%%%%%%%%%%%%%%%%%%%%%%%%%%%%%%%%%%%%%%%%%%%%%%%

Compared to these recent works \cite{comparisonPaper2017, Vourvoulakis, KAPELA} of implementation of matching cores on hardware, our proposed architecture consumes significantly fewer resources with acceptable matching accuracy (~98\%).
%%%%%%%%%%%%%%%%%%%%%%%%%%%%%%%%%%%%%%%%%%%%%%%%%%%%%%%%%

%%%%%%%%%%%%%%%%%%%%%%%%%%%%%%%%%%%%%%%%%%%%%%%%%%%%%%%%%
%%%%%%%%%%%%%%%%%%% Conclusions %%%%%%%%%%%%%%%%%%%%%%%%%
%%%%%%%%%%%%%%%%%%%%%%%%%%%%%%%%%%%%%%%%%%%%%%%%%%%%%%%%%
\section{Conclusions} \label{sec:conclusion}

In this paper, a fully pipelined accelerator for a keypoint descriptor matching scheme for the SIFT object recognition algorithm was designed and implemented on FPGA where the matching core was constructed of four main computational sub-modules and two local caches. Utilizing a close construction and 16-bit fixed-point calculations helped alleviate memory bandwidth restrictions in order to achieve maximum throughput. An experimental system was designed on a Xilinx\textsuperscript{\textregistered} Zedboard where the matching core was implemented on the programmable fabric and the Zynq processing system initialized the matching process. Our proposed SIFT matching architecture consumes fewer resources and accelerates the matching process where 9.11, 6.75, and 6.08 milliseconds elapsed for calculating the matching points of 882, 638, and 579 descriptors, respectively, with an image of 1021 descriptors. Our proposed SIFT matching hardware implementation additionally utilized 91\% fewer LUTs and 79\% fewer BRAM when comparing with the state of the art hardware matching core. Future work includes the extension of the hardware architecture into a fully pipelined vision system with increased number of computation cores.

\bibliographystyle{elsarticle-num}
\bibliography{main_paper}

\begin{thebibliography}{10}
\expandafter\ifx\csname url\endcsname\relax
  \def\url#1{\texttt{#1}}\fi
\expandafter\ifx\csname urlprefix\endcsname\relax\def\urlprefix{URL }\fi
\expandafter\ifx\csname href\endcsname\relax
  \def\href#1#2{#2} \def\path#1{#1}\fi

\bibitem{DavidLowe1999}
D.~G. Lowe, {Object Recognition from Local Scale-Invariant Features}, in:
  Proceedings of the Seventh IEEE International Conference on Computer Vision,
  Vol.~2, 1999, pp. 1150--1157.

\bibitem{MedApp1}
H.~B.-S. Lee D-H, Lee D-W, {Possibility Study of Scale Invariant Feature
  Transform (SIFT) Algorithm Application to Spine Magnetic Resonance Imaging},
  PLoS ONE 11~(4).

\bibitem{SatelliteApp2}
Y.~Jiang, Y.~Xu, Y.~Liu, Performance evaluation of feature detection and
  matching in stereo visual odometry, Neurocomputing 120 (2013) 380 -- 390,
  image Feature Detection and Description.

\bibitem{FacialRecog1}
J.~Križaj, V.~Štruc, N.~Pavešic, {Adaptation of SIFT Features for Face
  Recognition under Varying Illumination}, in: The 33rd International
  Convention MIPRO, 2010, pp. 691--694.

\bibitem{UVApp1}
A.~Cesetti, E.~Frontoni, A.~Mancini, A.~Ascani, P.~Zingaretti, S.~Longhi, {A
  Visual Global Positioning System for Unmanned Aerial Vehicles Used in
  Photogrammetric Applications}, Journal of Intelligent {\&} Robotic Systems
  61~(1) (2011) 157--168.

\bibitem{kolman2004elementary}
B.~Kolman, D.~R. Hill, {Elementary Linear Algebra}, Pearson Education, 2004.

\bibitem{Chi_square}
G.~Qian, S.~Sural, Y.~Gu, S.~Pramanik, {Similarity Between Euclidean and Cosine
  Angle Distance for Nearest Neighbor Queries}, in: Proceedings of the 2004 ACM
  Symposium on Applied Computing, SAC '04, ACM, New York, NY, USA, 2004, pp.
  1232--1237.

\bibitem{Vourvoulakis}
J.~Vourvoulakis, J.~Kalomiros, J.~Lygouras, Fpga-based architecture of a
  real-time sift matcher and ransac algorithm for robotic vision applications,
  Multimedia Tools and Applications 77~(8) (2018) 9393--9415.

\bibitem{comparisonPaper2017}
G.~Lentaris, I.~Stamoulias, D.~Soudris, M.~Lourakis, {HW/SW Codesign and FPGA
  Acceleration of Visual Odometry Algorithms for Rover Navigation on Mars},
  IEEE Transactions on Circuits and Systems for Video Technology 26~(8) (2016)
  1563--1577.

\bibitem{Wang}
J.~{Wang}, S.~{Zhong}, L.~{Yan}, Z.~{Cao}, {An Embedded System-on-Chip
  Architecture for Real-time Visual Detection and Matching}, IEEE Transactions
  on Circuits and Systems for Video Technology 24~(3) (2014) 525--538.

\bibitem{KAPELA}
R.~Kapela, K.~Gugala, P.~Sniatala, A.~Swietlicka, K.~Kolanowski, {Embedded
  platform for local image descriptor based object detection}, Applied
  Mathematics and Computation 267 (2015) 419 -- 426, the Fourth European
  Seminar on Computing (ESCO 2014).

\bibitem{BRIEF}
M.~{Calonder}, V.~{Lepetit}, M.~{Ozuysal}, T.~{Trzcinski}, C.~{Strecha},
  P.~{Fua}, {BRIEF: Computing a Local Binary Descriptor Very Fast}, IEEE
  Transactions on Pattern Analysis and Machine Intelligence 34~(7) (2012)
  1281--1298.

\bibitem{FREAK}
A.~{Alahi}, R.~{Ortiz}, P.~{Vandergheynst}, {FREAK: Fast Retina Keypoint}, in:
  2012 IEEE Conference on Computer Vision and Pattern Recognition, 2012, pp.
  510--517.

\bibitem{CONDELLO}
G.~Condello, P.~Pasteris, D.~Pau, M.~Sami, {An OpenCL-based feature matcher},
  Signal Processing: Image Communication 28~(4) (2013) 345 -- 350, special
  Issue: VS\&AR.

\bibitem{fassold2015gpu}
H.~Fassold, H.~Stiegler, J.~Rosner, M.~Thaler, W.~Bailer, {A GPU-accelerated
  two stage visual matching pipeline for image and video retrieval}, in:
  Content-Based Multimedia Indexing (CBMI), 2015 13th International Workshop
  on, IEEE, 2015, pp. 1--5.

\bibitem{zynqboardmanual}
{Xilinx\ Inc.}, {ZC702 Evaluation Board for the Zynq-7000 XC7Z020 User Guide}
  (September, 2015).

\bibitem{daoud2015survey}
L.~Daoud, D.~Zydek, H.~Selvaraj, {A Survey on Design and Implementation of
  Floating Point Adder in FPGA}, in: Progress in Systems Engineering, Springer,
  2015, pp. 885--892.

\bibitem{daoud2018sift}
L.~Daoud, M.~K. Latif, N.~Rafla, {SIFT Keypoint Descriptor Matching Algorithm:
  A Fully Pipelined Accelerator on FPGA}, in: Proceedings of the 2018 ACM/SIGDA
  International Symposium on Field-Programmable Gate Arrays, ACM, 2018, pp.
  294--294.

\bibitem{daoud2014survey}
{L. Daoud, D. Zydek, and H. Selvaraj}, {A Survey of High Level Synthesis
  Languages, Tools, and Compilers for Reconfigurable High Performance
  Computing}, in: Advances in Systems Science, Springer, 2014, pp. 483--492, ,
  DOI: 10.1007/978-3-319-01857-7\_47.

\bibitem{systemgenerator}
{{Xilinx} Inc.}, {Vivado Design Suite Reference Guide: Model-Based DSP Design
  Using System Generator} (May, 2019).

\bibitem{axidmamanual}
{Xilinx Inc.}, {AXI DMA v7.1: LogiCORE IP Product Guide} (October, 2017).

\end{thebibliography}

\end{document}